\documentclass[conference]{IEEEtran}
\IEEEoverridecommandlockouts
\usepackage{cite}
\usepackage{amsmath,amssymb,amsfonts}
\usepackage{algorithmic}
\usepackage{enumitem}
\usepackage{float}
\usepackage{multirow}
\usepackage{comment}

\usepackage{graphicx}
\usepackage{bm}         
\usepackage{booktabs}
\usepackage{array}      
\usepackage{subcaption} 
\usepackage{dblfloatfix} 
\usepackage{textcomp}
\usepackage{tablefootnote}
\usepackage{xcolor}

\def\BibTeX{{\rm B\kern-.05em{\sc i\kern-.025em b}\kern-.08em
    T\kern-.1667em\lower.7ex\hbox{E}\kern-.125emX}}
\begin{document}

\title{Uncertainty-Driven Anomaly Detection for Psychotic Relapse Using Smartwatches: \\ Forecasting and Multi-Task Learning Fusion}
\author{
\IEEEauthorblockN{
N.~Tsalkitzis$^{2}$,
P.~P.~Filntisis$^{1,3}$,
P.~Maragos$^{1,2,3}$
and N.~Efthymiou$^{1,3}$
}
\IEEEauthorblockA{$^{1}$Robotics Institute, Athena Research and Innovation Center, Greece}
\IEEEauthorblockA{$^{2}$School of ECE, National Technical University of Athens, Greece}
\IEEEauthorblockA{$^{3}$HERON - Hellenic Robotics Center of Excellence, Athens, Greece}
\IEEEauthorblockA{\texttt{\{nefthymiou, pflntisis\}@athenarc.gr, \{maragos\}@cs.ntua.gr, nikostsalkitzhs@gmail.com}}
}

\maketitle

\vspace{-0.6em}

\begin{abstract}
Digital phenotyping enables continuous passive monitoring of behavior and physiology, offering a promising paradigm for early detection of psychotic relapse. In this work, we develop and systematically study two smartwatch-based frameworks for daily relapse detection. The first forecasts cardiac dynamics and flags deviations between predicted and observed features as indicators of abnormality. The second adopts a multi-task formulation that fuses sleep with motion and cardiac-derived signals, learning time-aware embeddings and predicting measurement timing. Both pipelines use Transformer encoders and output a daily anomaly score, derived from predictive uncertainty estimated via an ensemble of multilayer perceptrons to improve robustness to real-world wearable variability. While each framework independently demonstrates strong predictive power, we show that they capture complementary physiological signatures. Consequently, we propose a late-fusion strategy that synergistically combines the anomaly signals from both architectures into a unified decision score. We benchmark our methodology on the 2nd e-Prevention Grand Challenge dataset, where our fused model achieves a 8\% relative improvement over the competition-winning baseline. Our results, supported by extensive ablation studies, suggest that the integration of diverse digital phenotypes, cardiac, motion, and sleep, is essential for the high-fidelity detection of psychotic relapse in real-world settings.


\end{abstract}


\section{Introduction}
\begin{figure*}[!t]
    \centering
    \includegraphics[width=\textwidth]
    {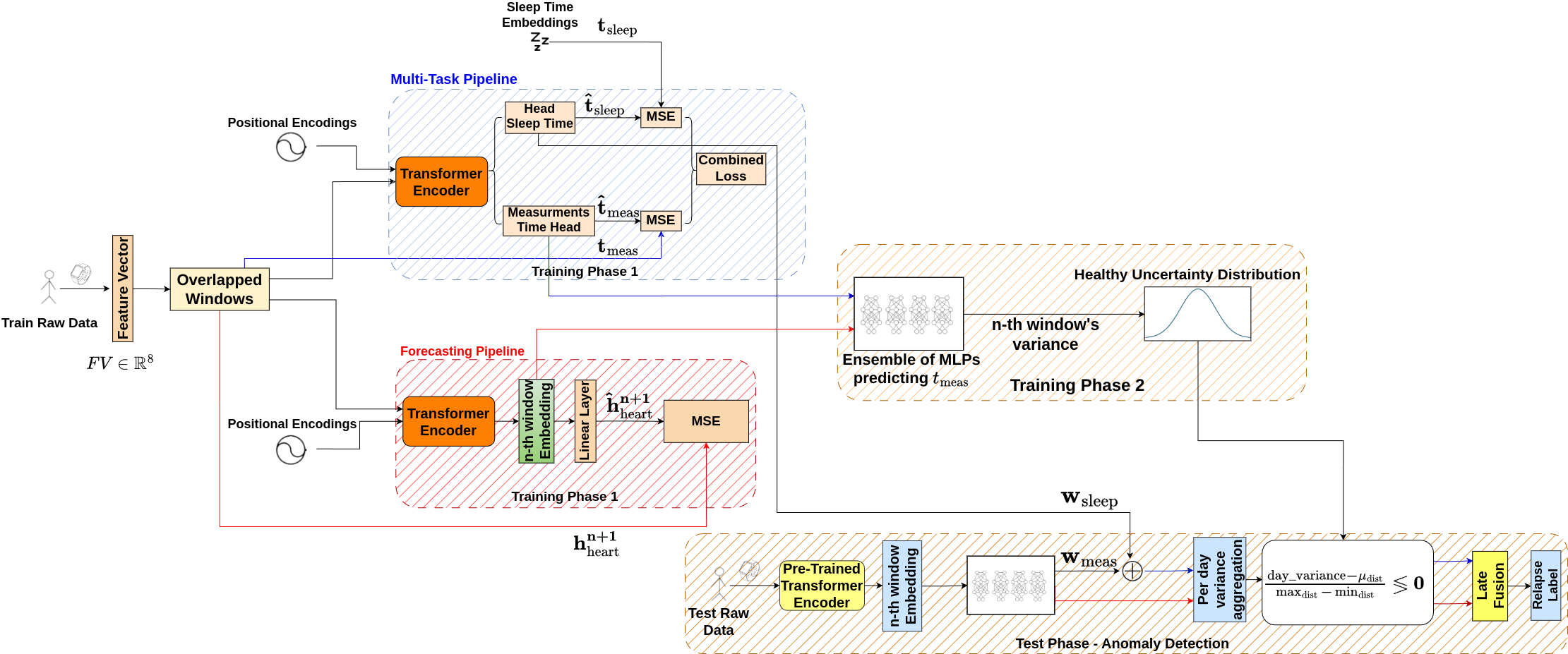}
    \caption{The proposed transformer-based relapse and anomaly detection framework where windowed wearable features are encoded and then routed through one of two alternative learning paths either a \textbf{forecasting} pipeline shown in red or a \textbf{multi-task} sleep and measurement time pipeline shown in blue. The resulting predictions are used to derive uncertainty estimates via an MLP ensemble which are aggregated and late fused during testing to produce the final relapse label.}

    \label{fig:architectures}
\end{figure*}

Digital phenotyping has increasingly emerged as a valuable approach in medical applications, with particularly strong momentum in psychiatry \cite{b1,b2}. By leveraging unobtrusively digital data during everyday life \cite{b3, b4}, clinicians can enrich standard assessments with additional context about an individual’s mental state. Because these measurements are collected continuously in naturalistic settings, they can reveal both overt behavioral changes and subtle shifts that may precede clinical deterioration \cite{b5}. This capability supports earlier detection of clinically meaningful changes, including signals of impending psychotic relapse, and allows for personalized interventions that augment traditional care \cite{b6}.

Relapse detection is typically framed as supervised learning when labels are abundant, or unsupervised when events are rare. While \cite{b7} uses a bidirectional LSTM for weekly probability estimates, the scarcity of real-world relapse data often favors unsupervised models that learn a ``remission baseline." For instance, \cite{b8} treats identity misclassification as a proxy for clinical change, while \cite{b9} uses a Convolutional Autoencoder to extract latent features from 4-hour 2D profiles, deriving relapse scores from sleep-specific reconstruction errors. More complex architectures include multi-branch CNNs with multi-head attention to visualize 24-hour activity patterns \cite{b10}, and Transformer-based encoder-decoders that leverage smartphone features to flag individualized behavioral anomalies \cite{b4}.

Recently, during the 2nd e-Prevention Grand Challenge on psychotic relapse detection a biosignal dataset \cite{b11} was introduced. SoTA work \cite{b12} in the challenge employs a patient-specific 
Transformer-based unsupervised anomaly detection framework that learns an individual’s typical daily routine by predicting the timestamp of wearable biosignal measurements using non-relapse data, deriving daily anomaly scores via an ensemble of MLP heads to stabilize prediction error. 
Other approaches, such as \cite{b13}, utilize separate autoencoders for multimodal data combined with Elliptic Envelopes to calculate relapse probabilities, while \cite{b14} leverages a Transformer encoder with cross-entropy and prototype losses to identify outlier days.


Building on these foundations, we present a novel Transformer-based framework for modeling wearable-derived physiological signals taking advantages of both cardiac variability and sleep-wake regularity that are associated with within-subject fluctuations in psychotic symptom severity, and departures from an individual's typical patterns may therefore provide early evidence of relapse risk \cite{b15}. Specifically, we study two approaches, one for forecasting future cardiac features and the second learning time-aware and sleep embeddings via a multi-task Transformer encoder, where an MLP ensemble estimates predictive uncertainty to derive the final anomaly score. Their fusion results in a powerful framework for enabling early detection of relapses.


In this paper, our contributions are summarized as follows:
\begin{itemize}
    \setlength{\leftmargin}{1.2em}
    \setlength{\itemsep}{0.15em}
    \setlength{\topsep}{0.2em}
    \setlength{\parsep}{0pt}
    \setlength{\partopsep}{0pt}   
\item We develop two smartwatch-based Transformer architectures for daily psychotic relapse detection: one for cardiac feature forecasting, and another that leverages multi-task learning with sleep-based, time-aware embeddings. Both models deliver strong performance on the e-Prevention benchmark, achieving a 7\% improvement over the SoTA work. 
\item We systematically investigate late-fusion strategies for combining the two models and show that jointly leveraging their anomaly scores yields improved performance, reaching an 8\% gain.

\end{itemize}

\section{Dataset}
For our study, we use the dataset from Track 2 of the ICASSP 2024 e-Prevention Grand Challenge \cite{b11}, part of the e-Prevention dataset \cite{b16}, which consists of smartwatch-derived biosignals from eight patients with psychotic disorders, such as schizophrenia and bipolar disorder, all of whom experienced at least one psychotic relapse during the recording period. The raw dataset comprises continuous recordings from multiple modalities, including accelerometer and gyroscope signals, RR intervals, heart rate, step counts, and sleep-related activity. For analysis, each participant’s recordings are segmented into daily windows, and the raw measurements are aggregated into 5-minute intervals to reduce the impact of sensor-level noise on downstream classification, consistent with the considerations reported in \cite{b17}. Subsequently, we carefully extract the following features: the acceleration norm and gyroscope norm, the mean heart rate, and heart-rate-variability descriptors computed from the RR-interval series, including the mean RR interval, RMSSD, SDNN, and the high-frequency Lomb-Scargle power. In addition, we incorporate time embeddings for the measurement timestamps, as well as embeddings for the sleep onset and wake-up times. 

On average, the training split contains approximately 200 days of recordings, while the validation and test splits comprise about 87 and 85 days, respectively. Importantly, the training data include remission periods only, whereas both the validation and test data contain samples from remission as well as relapse periods. After extracting features for each participant, we segment each day into overlapping windows formed over the 288 five-minute intervals. The number of windows per day is given by
\begin{equation}
N_{\text{win}} = \left\lfloor \frac{288-\emph {window\_size}}{\emph {stride}} \right\rfloor + 1,
\end{equation}
where 288 denotes the number of five-minute intervals in a day, \emph{stride} controls the displacement of the sliding window, and \emph{window\_size} denotes the window length.

\section{Framework Architecture}
The proposed architectures, both based on Transformer encoders, are presented in Fig.~\ref{fig:architectures}. In the forecasting pipeline, the encoder is used to predict future trajectories of cardiac features, whereas the multi-task model jointly learns time-aware and sleep-related embeddings. Subsequently, an ensemble of multilayer perceptrons is employed to quantify predictive uncertainty which is then aggregated into an uncertainty score and, ultimately, utilized to derive the final anomaly score. This approach leverages the fact that shifts in cardiac and sleep–wake patterns often track with psychotic symptom severity and provide early warnings of impending relapse.

\subsection{Forecasting Architecture}
For \textbf{the forecasting-based pipeline}, the input at each five-minute timestep consists of the acceleration and gyroscope norm, the five cardiac features and the count of steps. These sequences are passed through a Transformer encoder, which produces a latent representation per timestep. Average pooling over the window yields a single embedding per window, which is subsequently used to predict the cardiac-feature vector at the next timestep.
Formally, let $\mathbf{x}_t \in \mathbb{R}^{d}$ denote the input feature vector at timestep $t$, and let $\mathbf{c}_t \in \mathbb{R}^{5}$ denote the corresponding vector of cardiac features. For a window of length $L$ starting at index $s$, define the input segment $\mathbf{X}_s = [\mathbf{x}_s,\ldots,\mathbf{x}_{s+L-1}]$. Given $\mathbf{X}_s$, the encoder produces a sequence of hidden states, 
$\mathbf{H}_s = \mathrm{Enc}(\mathbf{X}_s),$
where $\mathbf{H}_s = [\mathbf{H}_{s,0}, \ldots, \mathbf{H}_{s,L-1}]$ and $\mathbf{H}_{s,i}$ denotes the encoder output at offset $i$ within the window. We obtain a fixed-dimensional representation for the entire window by mean pooling,
\vspace{-0.2cm}
\begin{align}
\mathbf{z}_s &= \frac{1}{L}\sum_{i=0}^{L-1}\mathbf{H}_{s,i},
\end{align}
and pass this window-level embedding to a predictor head $\mathbf{g}$ to forecast the next-step cardiac features,
$\hat{\mathbf{c}}_{s+L} = g(\mathbf{z}_s)$.
The model is trained using a one-step-ahead forecasting objective over a batch $\mathcal{B}$ of windows,
\begin{equation}
\mathcal{L}_{\mathrm{forecast}} =
\frac{1}{\mathcal{B}}\sum_{s\in\mathcal{B}}
\left\| \hat{\mathbf{c}}_{s+L} - \mathbf{c}_{s+L} \right\|_2^2.
\end{equation}
\vspace{-0.2cm}

In a second training phase, we reuse the same encoder embedding $\mathbf{z}_s$, but replace the single predictor $\mathbf{g}$ with an ensemble of $K$ multilayer perceptrons (MLPs). The learning objective remains the same one-step-ahead cardiac forecasting loss, while diversity is promoted via an online resampling scheme. For each mini-batch, embedding-target pairs are replicated across all $K$ heads, and then, independently per head, a subset of samples is replaced with embedding-target pairs drawn from randomly selected training instances. Let $\{f_k(\cdot)\}_{k=1}^{K}$ denote the ensemble and let $\tilde{\mathbf{z}}_{s}^{(k)}$ denote the resampled input to head $k$. Each head produces a next-step prediction,
\begin{equation}
\hat{\mathbf{c}}_{s+L}^{(k)} = f_k\!\left(\tilde{\mathbf{z}}_{s}^{(k)}\right), \qquad k=1,\ldots,K,
\end{equation}
and we aggregate predictions using the ensemble mean,
\begin{equation}
\bar{\mathbf{c}}_{s+L} = \frac{1}{K}\sum_{k=1}^{K}\hat{\mathbf{c}}_{s+L}^{(k)}.
\end{equation}

To quantify predictive uncertainty for window $s$, we compute the element-wise ensemble variance,
\begin{equation}
\boldsymbol{\sigma}^2_{s} =
\frac{1}{K}\sum_{k=1}^{K}\left(\hat{\mathbf{c}}_{s+L}^{(k)}-\bar{\mathbf{c}}_{s+L}\right)^{\odot 2},
\end{equation}
where $(\cdot)^{\odot 2}$ denotes element-wise squaring and $\boldsymbol{\sigma}^2_{s,j}$ corresponds to the variance of the $j$-th cardiac feature. We summarize this vector by averaging across the five cardiac-feature dimensions, \vspace{-0.4cm}
\begin{equation}
u_s = \frac{1}{5}\sum_{j=1}^{5}\boldsymbol{\sigma}^2_{s,j}.
\end{equation}
Finally, the daily variance score is defined as the mean uncertainty across all windows whose start indices fall within day $d$,
\begin{equation}
U_d = \frac{1}{\mathcal{W}_d}\sum_{s\in \mathcal{W}_d} u_s,
\end{equation}
where $\mathcal{W}_d$ representing the number of windows in day $d$.

Using the remission-only training split, we construct a patient-specific (healthy) uncertainty distribution from the day-level scores $\{U_d\}$ and summarize it via its empirical minimum, maximum, and mean values. During inference, the proposed architecture produces a day-level uncertainty score $U_d$ for each day $d$. For remission days, this score is expected to remain close to the mean of the healthy distribution, whereas relapse days are characterized by systematic deviations. Accordingly, we define the final anomaly score for day $d$ as the normalized deviation from the healthy mean:
\begin{equation}
A_d = \frac{U_d - \mu_{\text{dist}}}{\max_{\text{dist}} - \min_{\text{dist}}},
\end{equation}
where $\mu_{\text{dist}}$, $\max_{\text{dist}}$, and $\min_{\text{dist}}$ denote the mean, maximum, and minimum of the patient-specific healthy uncertainty distribution, respectively. Finally, we apply a fixed decision rule by thresholding the anomaly score:
\begin{equation}
\hat{y}_d =
\begin{cases}
1, & \text{if } A_d > \tau \quad (\text{relapse}),\\
0, & \text{otherwise} \quad (\text{remission}).
\end{cases}
\end{equation}
\subsection{Multi-Task Architecture}
In contrast, for \textbf{the multi-task learning approach} we build upon and extend the winning architecture of the challenge \cite{b12}. The model receives two input streams. The first comprises physiological and activity-related signals, including cardiac features, the norms of the inertial sensors, and step counts. The second stream consists of two pairs of time embeddings that encode sleep onset and wake-up time. Training is performed with a multi-task objective that combines (i) the prediction of measurement-time embeddings from the observed signals and (ii) an auxiliary sleep-prediction head that estimates sleep-related time embeddings. For uncertainty estimation and the construction of the patient-specific healthy distribution, we exclude the sleep branch and rely solely on uncertainty derived from the measurement-time embedding task, since sleep-related measurements are comparatively noisy in the available data\cite{b12}. During inference, we compute a combined day-level variance score by fusing the variance from the sleep head with the uncertainty associated with the measurement-time embeddings, using weights of 0.3 and 0.7, respectively. The resulting score is then normalized using the summary statistics of the healthy distribution to obtain the final anomaly score.
\subsection{Late Fusion} To leverage the complementary strengths of the heart-forecasting and the sleep/circadian architectures, we investigated \emph{late fusion} at the score level. Each model produces a continuous day-level anomaly score and instead of thresholding the two systems independently, we fuse these continuous scores into a single risk score and apply thresholding only at the end. This retains more information for ranking-based evaluation and allows the two modalities to compensate for one another. Our primary fusion mechanism is a convex combination of the two anomaly scores:
\begin{equation}
\text{fused\_score}
=
\alpha \cdot A_{D,\text{heart}}
+
(1-\alpha)\cdot A_{D,\text{sleep}}
\end{equation}

where $\alpha \in [0,1]$ controls the relative contribution of each modality. The value of $\alpha$ was selected on the validation set,via grid search, to maximize detection performance. In addition to weighted averaging, we explored max and min fusion:
\begin{equation}
\text{fused\_score} = \{\max,\min\}(A_{D,\text{heart}},\, A_{D,\text{sleep}})
\end{equation}

Max fusion acts as an \emph{OR}-style rule, producing a high fused score when \emph{either} model strongly signals abnormality. This is beneficial when relapses may manifest predominantly in one modality (e.g., sleep disruption without marked cardiac changes, or vice versa), thereby improving sensitivity. Conversely, min fusion behaves like an \emph{AND}-style rule, yielding a high fused score only when \emph{both} modalities are simultaneously anomalous. This can reduce false positives by requiring cross-modal agreement, at the cost of lower recall when a relapse signal is strong in only one of the two models.
\section{Experimental Analysis}
All models were trained for 50 epochs with a learning rate of $10^{-3}$, weight decay $5\times10^{-4}$, and a batch size of 16. We used an ensemble of five MLP predictors. Performance was evaluated using AUROC, AUPRC, and their arithmetic mean (AVG), reported as the average over 10 independent runs.

\textbf{Positional Embeddings:} Since all methods are Transformer-based we compared positional encodings per task. Table \ref{tab:posenc_results} showcases that ALiBi \cite{b18} is most effective for sleep multi‑task modeling because its decaying attention bias, where scores are penalized by token distance, inherently suits sleep architecture. This method captures local context such as stage transitions while staying invariant to absolute shifts between nights, matching the relative and variable nature of sleep recordings. For cardiac forecasting sinusoidal embeddings perform best by providing an absolute temporal anchor that preserves consistent periodic signals like circadian and time‑of‑day effects. RoPE \cite{b19} is competitive but its rotation‑based relative encoding yields lower performance than the best method for each task.

\begin{table}[!htbp]
\centering
\small
\setlength{\tabcolsep}{6pt}
\renewcommand{\arraystretch}{1.15}
\caption{Positional Embedding Performance. Best per model/metric in \textbf{bold}; second-best is underlined.}
\label{tab:posenc_results}
\resizebox{\columnwidth}{!}{%
\begin{tabular}{llccc}
\toprule
Model & Positional encoding & AUROC & AUPRC & AVG \\
\midrule
\multirow{3}{*}{Forecasting}
& RoPE\cite{b19}       & \underline{0.562} & \underline{0.549} & \underline{0.556} \\
& ALiBi\cite{b18}      & 0.547 & 0.531 & 0.539 \\
& sinusoidal           & \textbf{0.565} & \textbf{0.552} & \textbf{0.559} \\
\midrule
\multirow{3}{*}{Multi-task}
& RoPE\cite{b19}       & \underline{0.505} & \underline{0.608} & \underline{0.557} \\
& ALiBi\cite{b18}      & \textbf{0.509} & \textbf{0.612} & \textbf{0.561} \\
& sinusoidal           & 0.503 & \underline{0.608} & 0.556 \\
\bottomrule
\end{tabular}%
}
\end{table}

\noindent\textbf{Window Hyperparameters Ablation:} The number of windows is governed by two key parameters, namely the stride and the window size. To determine an appropriate stride value for both models, we conducted a sensitivity analysis, the results of which are reported in Table~\ref{tab:stride_results}.

\begin{table}[!htbp]
\centering
\small
\setlength{\tabcolsep}{4pt}
\renewcommand{\arraystretch}{1.15}
\caption{Stride sensitivity (mean $\pm$ std). Best per model/metric in \textbf{bold}; second-best is underlined.}
\label{tab:stride_results}
\resizebox{\columnwidth}{!}{%
\begin{tabular}{llccc}
\hline
Model & Stride & AUROC & AUPRC & AVG \\
\hline
\multirow{4}{*}{Forecasting}
& 12 (1 hr) & \textbf{0.568 $\pm$ 0.022} & \textbf{0.554 $\pm$ 0.019} & \textbf{0.561 $\pm$ 0.019} \\
& 24 (2 hr) & 0.538 $\pm$ 0.020 & \underline{0.551 $\pm$ 0.015} & 0.545 $\pm$ 0.017 \\
& 36 (3 hr) & \underline{0.542 $\pm$ 0.019} & 0.547 $\pm$ 0.019 & \underline{0.545 $\pm$ 0.018} \\
& 48 (4 hr) & 0.535 $\pm$ 0.029 & 0.522 $\pm$ 0.027 & 0.528 $\pm$ 0.028 \\
\hline
\multirow{4}{*}{Multi-task}
& 12 (1 hr) & \underline{0.506 $\pm$ 0.009} & \underline{0.615 $\pm$ 0.011} & \underline{0.560 $\pm$ 0.009} \\
& 24 (2 hr) & \textbf{0.506 $\pm$ 0.005} & \textbf{0.617 $\pm$ 0.008} & \textbf{0.562 $\pm$ 0.006} \\
& 36 (3 hr) & 0.503 $\pm$ 0.013 & 0.594 $\pm$ 0.024 & 0.548 $\pm$ 0.017 \\
& 48 (4 hr) & 0.496 $\pm$ 0.010 & 0.574 $\pm$ 0.025 & 0.535 $\pm$ 0.016 \\
\hline
\end{tabular}%
}
\end{table}

With the stride results above, a one-hour stride yields the best performance for both architectures. Keeping this stride fixed, a subsequent window-size sensitivity analysis indicates that a two-hour window (24 timesteps) is optimal, corresponding to a 50\% overlap between consecutive windows. This choice provides a favorable trade-off between data efficiency and temporal context: it increases the number of training samples while capturing meaningful short-term dynamics, without overly long windows that may dilute relapse-related patterns and introduce additional noise or nonstationarities.

\textbf{Threshold $\tau$ Ablation:} We conducted an ablation study on the validation set to examine how the relapse decision threshold $\tau$ affects performance. This threshold acts as a decision cutoff. It represents the minimum daily risk score required for a day to be labeled as a relapse. Decreasing $\tau$ makes relapse detection more permissive, whereas increasing $\tau$ makes it more conservative. Across both modalities, we observed that negative thresholds generally provide a better balance between ranking ability and precision under class imbalance. In addition, the two models show distinct strengths. The forecasting model tends to achieve stronger discrimination (higher AUROC), while the multi-task model typically yields better early-retrieval behavior (higher AUPRC), with this contrast being especially evident around $\tau=0$. Therefore, we select $\tau=-0.1$ and $\tau=-0.2$, respectively for the two architectures.

\section{Results \& Discussion} 
\begin{figure}[!htbp]
\centering
\includegraphics[width=\columnwidth,height=1\textheight,keepaspectratio]{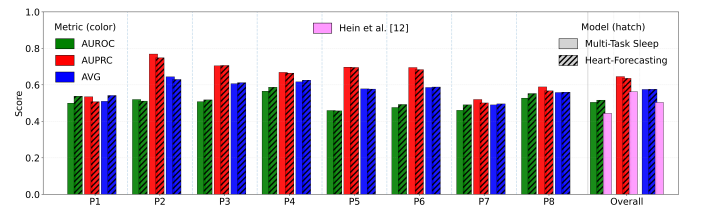}
\captionsetup{skip=2pt}
\caption{Patient-wise comparison of forecasting and multi-task models against Hein \textit{et al.}~[12] (overall baseline).}
\label{fig:placeholder}
\end{figure}

To assess which model yields the strongest performance, we further benchmark our methods against both the baseline and first two winning architectures of the challenge, with the comparative results summarized in Table~\ref{tab:all_compact} and Figure \ref{fig:placeholder}. 

\begin{table}[H]
\centering
\small
\setlength{\tabcolsep}{4pt}
\renewcommand{\arraystretch}{1.05}
\caption{Test-set performance for Track 2 reference/proposed models and score-fusion variants (best in \textbf{bold}; second best underlined).}
\label{tab:all_compact}
\begin{tabular}{llccc}
\toprule
\textbf{Group} & \textbf{Method} & \textbf{AUROC} & \textbf{AUPRC} & \textbf{AVG} \\
\midrule
\multirow{5}{*}{Single} 
 & Baseline Approach & 0.495 & 0.379 & 0.437 \\
 & Hein \cite{b12}  & 0.444 & 0.563 & 0.504 \\
 & Mallo-Ragolta \cite{b13}  & 0.493 & 0.505 & 0.499 \\
 & Multi-task Model & \underline{0.505} & \textbf{0.646} & \underline{0.575} \\
 & Forecasting-based Model & \textbf{0.516} & \underline{0.636} & \textbf{0.576} \\
\midrule
\multirow{3}{*}{Fusion}

 & Weighted ($\alpha{=}0.7,\ \tau{=}{-}0.1$) & \underline{0.504} & \underline{0.664} & \textbf{0.584} \\
 & Max ($\tau{=}{-}0.1$) & 0.501 & \textbf{0.667} & \textbf{0.584} \\
 & Min ($\tau{=}{-}0.1$) & \textbf{0.510} & 0.653 & 0.581 \\
\bottomrule
\end{tabular}
\end{table}
Based on the results above, both proposed models outperform the 2nd challenge’s winning architecture of Hein \textit{et al.}~[17] by approximately 7\% in AVG, supporting the benefit of uncertainty-driven anomaly scoring.

Per-person analysis further confirms that the two architectures are functionally complementary. The sleep-based model more often achieves higher AUPRC (e.g., P2 and P4--P8), suggesting improved identification of rare relapse days, while the heart-based model more frequently yields higher AUROC (e.g., P3, P4, and P6--P8), indicating stronger overall separability between relapse and non-relapse days. This distinction is clinically relevant under pronounced class imbalance: AUPRC is typically more informative because it emphasizes performance on the relapse class. The heterogeneity in per-patient AVG (sleep higher for P2 and P5; heart higher for P3, P4, and P6--P8) further suggests that relapse signatures are patient-specific and modality-dependent, motivating individualized balancing of modalities.

Fusion improves robustness by leveraging strengths from both pipelines. Across fusion rules (with $\alpha$ and $\tau$ selected on the validation set), max fusion attains the highest AUPRC and ties for the best AVG, while weighted fusion matches the top AVG with slightly lower AUPRC. In contrast, min fusion achieves the highest AUROC but the lowest AUPRC, consistent with better global separability yet fewer high-confidence relapse predictions and reduced recall. Overall, late fusion provides the strongest final scheme, and patient-specific fusion weights may yield further gains.

\section{Conclusion}
We develop two personalized, unsupervised relapse detectors from smartwatch digital phenotypes that compute day-level anomaly scores via ensemble-based predictive uncertainty: a cardiac forecasting model and a sleep/time multi-task model. Both models surpass the challenge-winning baseline, capturing distinct aspects of relapse risk (cardiac dynamics primarily improve separability, while sleep/time structure is especially informative under severe class imbalance). Late score-level fusion integrates these signals into our final scheme and achieves the strongest overall results (AVG = 0.584; AUPRC up to 0.667), corresponding to an $\sim$8\% relative improvement over the winner. Future work will explore early fusion approaches that jointly model cardiac and time/sleep patterns within a unified network to learn cross-modal interactions that may further improve relapse detection.

\end{document}